\pdfoutput=1

\documentclass[11pt]{article}
\usepackage{graphicx}
\usepackage{multirow, makecell}
\usepackage{acl}
\usepackage{times}
\usepackage{latexsym}
\usepackage[T1]{fontenc}
\usepackage[utf8]{inputenc}
\usepackage{microtype}

\title{ExtraPhrase: Efficient Data Augmentation for Abstractive Summarization}

\author{Mengsay Loem, Sho Takase, Masahiro Kaneko, Naoaki Okazaki \\
         Tokyo Institute of Technology \\ 
         \{mengsay.loem, sho.takase, masahiro.kaneko\}@nlp.c.titech.ac.jp\\
         okazaki@c.titech.ac.jp}

\begin{document}
\maketitle
\begin{abstract}
Neural models trained with large amount of parallel data have achieved impressive performance in abstractive summarization tasks.
However, large-scale parallel corpora are expensive and challenging to construct.
In this work, we introduce a low-cost and effective strategy, \textbf{ExtraPhrase}, to augment training data for abstractive summarization tasks. 
ExtraPhrase constructs pseudo training data in two steps: extractive summarization and paraphrasing.
We extract major parts of an input text in the extractive summarization step, and obtain its diverse expressions with the paraphrasing step.
Through experiments, we show that ExtraPhrase improves the performance of abstractive summarization tasks by more than 0.50 points in ROUGE scores compared to the setting without data augmentation.
ExtraPhrase also outperforms existing methods such as back-translation and self-training.
We also show that ExtraPhrase is significantly effective when the amount of genuine training data is remarkably small, i.e., a low-resource setting.
Moreover, ExtraPhrase is more cost-efficient than the existing approaches.
\end{abstract}

\section{Introduction}
\label{sec:intro}

\begin{figure*}[t]
\centering
\includegraphics[width=\textwidth]{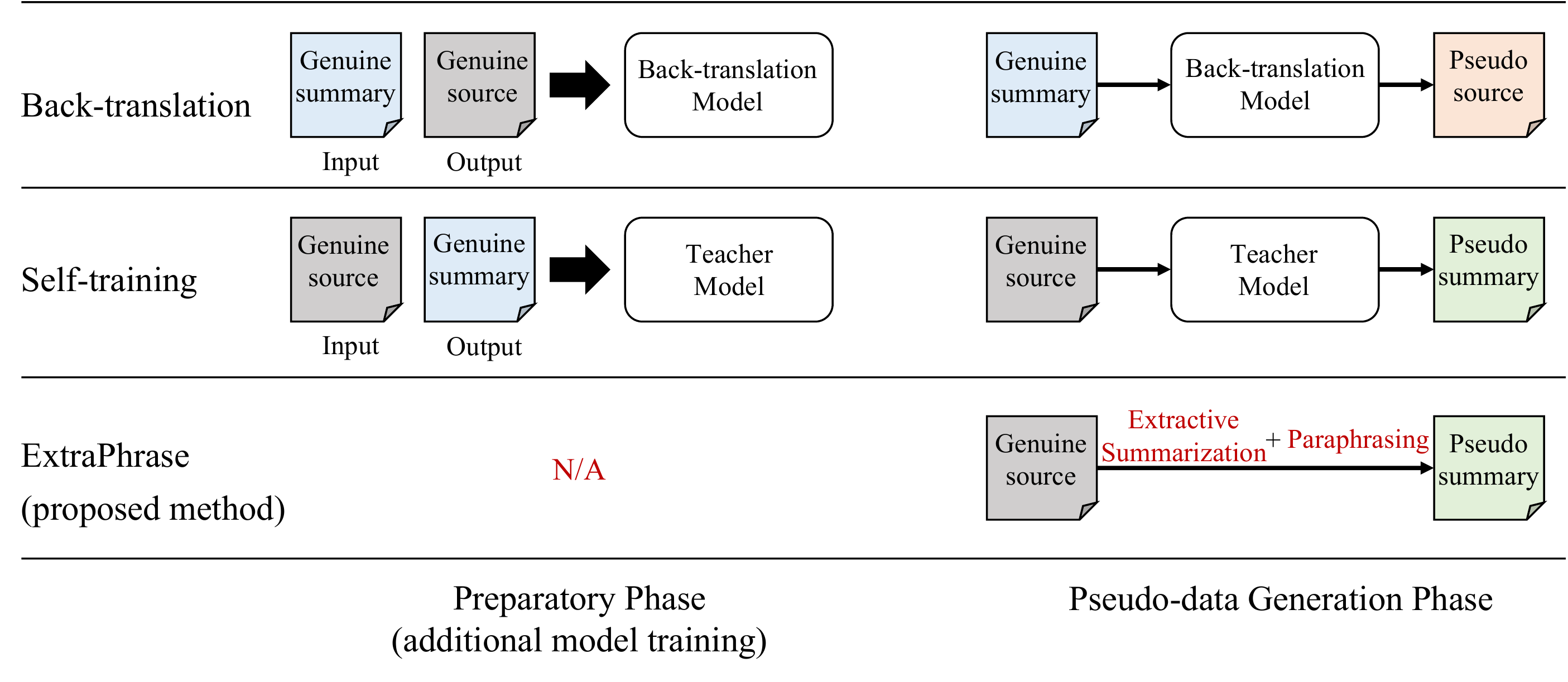}
\caption{\label{fig:related-work-method}Overview of pseudo data construction methods: back-translation, self-training, and \textbf{ExtraPhrase}.}
\end{figure*}

Neural encoder-decoders have achieved remarkable performance in various sequence-to-sequence tasks including machine translation, summarization, and grammatical error correction~\cite{DBLP:journals/corr/BahdanauCB14,rush-etal-2015-neural,yuan-briscoe-2016-grammatical}.
Recent studies indicated that neural methods are governed by the scaling law for the amount of training data~\cite{koehn-knowles-2017-six,NEURIPS2020_1457c0d6}.
In short, the more training data we prepare, the better performance a neural model achieves.
In this paper, we address increasing the training data for summarization to improve the performance of neural encoder-decoders on the abstractive summarization task.

In sequence-to-sequence tasks, we need a parallel corpus to train neural encoder-decoders.
Since it is too costly to construct genuine (i.e., human-generated) parallel corpora, most studies explored the way to construct pseudo training data automatically.
Back-translation is a widely used approach to construct pseudo training data for sequence-to-sequence tasks~\cite{sennrich-etal-2016-improving,edunov-etal-2018-understanding,caswell-etal-2019-tagged}.
In the back-translation approach, we construct a model generating a source side sentence from a target side sentence, and apply the model to a target side corpus to generate a pseudo source side corpus.
For example, in the English-to-German translation task, we construct a model generating English sentences from German sentences, and make the model generate pseudo English sentences from genuine German sentences.
In addition to machine translation, back-translation is also used in grammatical error correction~\cite{kiyono-etal-2019-empirical} and summarization~\cite{parida-motlicek-2019-abstract} tasks.
However, back-translation on summarization is an impossible problem essentially because a model is required to generate a source document (or sentence) from a summary.
In other words, the model is required to restore deleted information in the given summary without any guide.

\newcite{He2020Revisiting} indicated that the self-training approach, which makes a model generate target sentences from source sentences and use the pairs to train a model, can improve the performance on machine translation and summarization.
In contrast to back-translation, pseudo data generation for summarization by self-training is solvable but self-training is hard to generate diverse summaries \cite{gu2018non}.
Moreover, self-training and back-translation approaches require expensive computational cost because we need to train additional neural encoder-decoders on a large amount of training data to obtain high quality pseudo data~\cite{imankulova2019filtered}.

To solve these issues, we propose a novel strategy: \textbf{ExtraPhrase} consisting of \textbf{extra}ctive summarization and para\textbf{phrase} to construct pseudo training data for abstractive summarization.
Firstly, ExtraPhrase extracts an important part from a source text as a summary without requiring additional model training.
Then, we apply a paraphrasing technique to the extracted text to obtain diverse pseudo summaries.
Figure \ref{fig:related-work-method} illustrates comparison of ExtraPhrase and existing methods.

We conduct experiments on two summarization tasks: headline generation and document summarization tasks.
Experimental results show that pseudo training data constructed by our proposed strategy improves the performance on both tasks.
In detail, the pseudo data raises more than 0.50 in ROUGE F1 scores on both tasks.
Moreover, we conduct various analyses to explore the effectiveness and efficiency of ExtraPhrase.
In particular, we indicate that 1. ExtraPhrase is robust on low-resource settings (e.g., 1K genuine instances) and 2. ExtraPhrase is much more cost-efficient than previous self-training and back-translation approaches because ExtraPhrase does not require training additional neural encoder-decoders.

Our contributions are as follows:
\begin{itemize}
    \item We introduce ExtraPhrase consisting of extractive summarization and paraphrasing to construct pseudo summaries.
    \item We empirically show that pseudo training data constructed by ExtraPhrase improves the performance of neural encoder-decoders on abstractive summarization.
    \item We conduct various analyses to explore the property of our ExtraPhrase.
\end{itemize}

\section{Proposed Method: ExtraPhrase}
\label{sec:proposed_method}

As described in Section \ref{sec:intro}, our ExtraPhrase consists of two steps: extractive summarization and paraphrasing.
ExtraPhrase receives a (genuine) sentence as an input, and generates a pseudo summary corresponding to the input sentence.
When we construct a pseudo summary from a document, we independently apply ExtraPhrase to multiple sentences included in the given document.

\begin{figure*}[t]
\centering
\includegraphics[width=\textwidth]{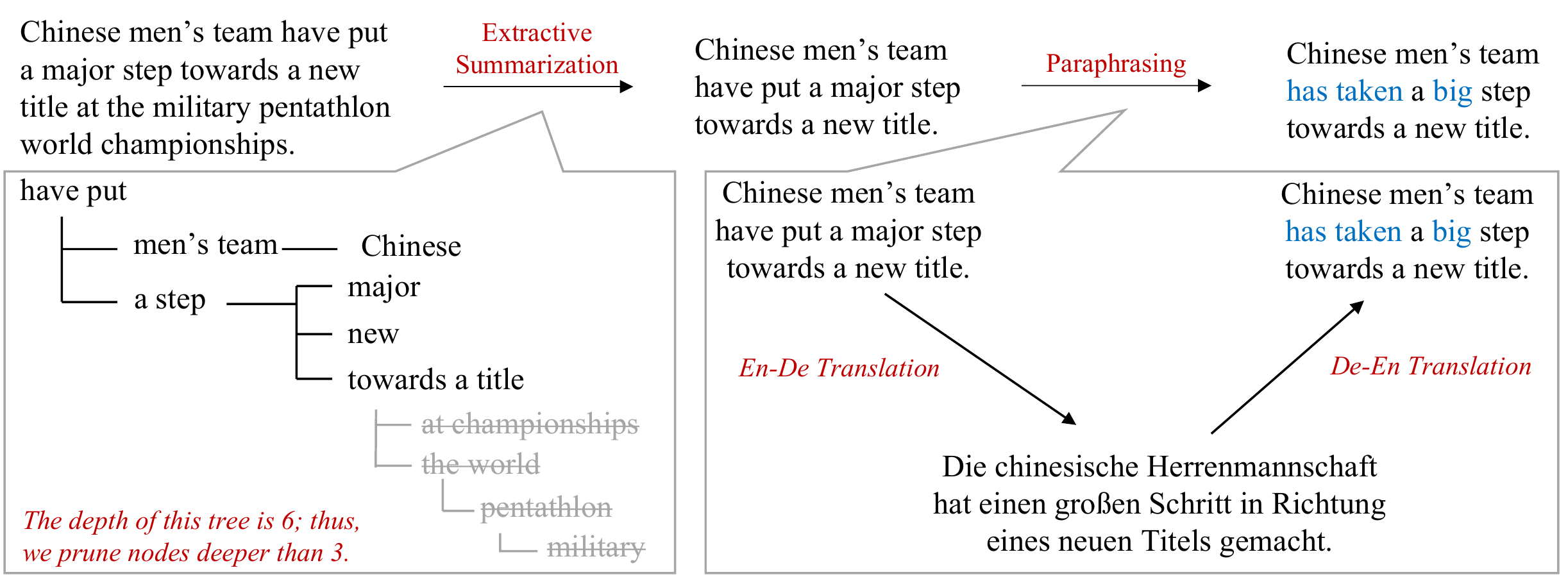}
\caption{\label{fig:proposed-method}Example of pseudo summary generated by \textbf{ExtraPhrase}. The upper part shows output sentences in each step of ExtraPhrase. Paraphrased words after paraphrasing (round-trip translation) in step-2 are highlighted in blue. The lower part shows the dependency tree of the source sentence omitted the dependency labels in step-1(left) and round-trip translation output in step-2 (right).}
\end{figure*}

Figure \ref{fig:proposed-method} illustrates the overview of ExtraPhrase briefly.
In the extractive summarization step, we extract important parts of a given sentence based on its dependency tree.
Then, we obtain another expression of the extracted sequence while keeping its meaning in the paraphrasing step.
We describe more details in this section.

\subsection{Step-1: Extractive Summarization}
\label{subsec:step-1}
In this extractive summarization step, we extract important parts of a given source sentence.
In other words, we conduct the sentence compression in this step.
Previous studies proposed various sentence compression methods such as rule-based methods~\cite{dorr-etal-2003-hedge}, the approach detecting important parts in a syntax tree~\cite{turner-charniak-2005-supervised,filippova-altun-2013-overcoming,DBLP:journals/corr/CohnL14}, sequential labeling approach~\cite{hirao-etal-2009-syntax}, and neural-based methods~\cite{filippova-etal-2015-sentence,kamigaito-etal-2018-higher}.

In this study, we adopt the simplest approach: a rule-based method based on the syntax tree of the given sentence.
Because the rule-based approach does not require any training corpus, we can use it in the situation where we do not have genuine parallel corpus.
We emphasize that we can use more sophisticated way if we need because we do not have any restriction for the summarization method in this step.

We define a rooted subtree of the syntax tree for the given sentence as important parts of the sentence.
We describe the procedure to extract such subtree.
First, we parse the given sentence to obtain its dependency tree.
Since the conventional parser assigns dependency relationships among words, each node in the dependency tree corresponds to a word.
However, functional words such as preposition and determiner are inappropriate granularity as nodes in deciding whether each node is important or not.
Thus, we combine such functional words with heads on the dependency tree as in \cite{filippova-altun-2013-overcoming}.

Then, we prune the dependency tree to obtain the small rooted subtree.
We can roughly select the output summary length (the number of words) by the depth of the subtree.
In this study, we prune nodes whose depth are deeper than half of the dependency tree.
The lower part of Figure \ref{fig:proposed-method} illustrates these processes.
For the dependency tree from the conventional parser (left side), we modify nodes consisting of functional words and prune nodes that are deeper than 3 (right side).

Finally, we linearize the extracted rooted subtree to obtain its sequential representation.
We follow the word order of the original sentence.
After this linearization, we obtain the summary of the given sentence.
To construct a summary of a document, we apply the above procedure to each sentence one-by-one.

\subsection{Step-2: Paraphrasing}
The constructed summaries by the previous step consist of words included in the source sentences only.
To increase the diversity of the summaries, we apply the paraphrasing method to the summaries.
For paraphrasing, we adopt the approach using machine translation models~\cite{sun-zhou-2012-joint,mallinson-etal-2017-paraphrasing} because some studies published high-quality neural machine translation models~\cite{ott-etal-2018-scaling,ng-etal-2019-facebook}.
In this approach, we obtain paraphrases by conducting round-trip translation that translates a sentence into a different language and the translated sentence into the original language.
Since this approach can generate various paraphrases, some studies use the approach as data augmentation~\cite{wei2018fast}.

Thus, in this paraphrasing step, we apply a publicly available neural machine translation model to an extracted summary.
Then, we convert the translated sentence into the original language by a reverse translation model.
In the same as in the extractive summarization step, we input a sentence to translation models.
Therefore, we generate a paraphrase for each sentence one-by-one if the summary contains multiple sentences.

\section{Experiments}
\label{sec:exp}
To investigate the effect of ExtraPhrase, we conduct experiments on two summarization tasks: headline generation and document summarization tasks.

\begin{table}
\centering
\begin{tabular}{lrr}
\hline
Dataset & Gigaword & CNN/DailyMail\\
\hline
Training & 3.8M & 287K\\
Validation & 3.0K & 13K\\
Test & 1.9K & 11K\\\hline
\end{tabular}
\caption{Statistic of datasets used in experiments}
\label{tab:datasets}
\end{table}

\begin{table*}
\centering
\begin{tabular}{lrrrrr}
\hline
Method & \#Genuine & \#Pseudo & ROUGE-1 & ROUGE-2 & ROUGE-L \\
\hline
Genuine only & 3.8M & -- & 37.95 & 18.80 & 35.05 \\
Oversampling & 7.6M & -- & 38.26 & 19.14 & 35.41 \\
\hline
Back-translation & 3.8M & 3.8M & 38.49 & 19.24 & 35.63 \\
Self-training & 3.8M & 3.8M & 38.32 & 19.06 & 35.37 \\
\hline

\textbf{ExtraPhrase} & 3.8M & 3.8M & \textbf{38.51} & \textbf{19.52} & \textbf{35.72} \\
\hline
\end{tabular}
\caption{\label{gigaword-rouge}
F1 based ROUGE scores for the headline generation task.
}
\end{table*}

\begin{table*}
\centering
\begin{tabular}{lrrrrr}
\hline
Method & \#Genuine & \#Pseudo & ROUGE-1 & ROUGE-2 & ROUGE-L \\
\hline
Genuine only & 287K & -- & 39.76 & 17.55 & 36.75 \\
Oversampling & 574K & -- & 40.14 & 17.86 & 37.05 \\
\hline
Back-translation & 287K & 287K & 39.93 & 17.74 & 36.85 \\
Self-training & 287K & 287K & 40.19 & 17.87 & 37.21 \\
\hline
\textbf{ExtraPhrase} & 287K & 287K & \textbf{40.57}& \textbf{18.22} & \textbf{37.51} \\
\hline
\end{tabular}
\caption{\label{cnndm-rouge}
F1 based ROUGE scores for the document summarization task.
}
\end{table*}

\subsection{Datasets}
\label{sec:dataset}
Table~\ref{tab:datasets} summarizes the genuine datasets used in this study.
The details are as below.

\paragraph{Headline generation}
We use the de-facto headline generation dataset constructed by \newcite{rush-etal-2015-neural}.
The dataset contains pairs of the first sentence and headline extracted from the annotated English Gigaword~\cite{napoles-etal-2012-annotated}.
We use the same splits for train, valid, and test as \newcite{rush-etal-2015-neural}.
We use the byte pair encoding~\cite{sennrich-etal-2016-neural} to construct a vocabulary set with the size of 32K by sharing vocabulary between source and target sides.

\paragraph{Document summarization}
CNN/DailyMail dataset~\cite{see-etal-2017-get} is widely used for the single document summarization task.
The training set contains 280K pairs of news articles and abstractive summary extracted from CNN and DailyMail websites.
In the same as the headline generation, we construct a vocabulary set with the byte pair encoding~\cite{sennrich-etal-2016-neural}.
We set the vocabulary size 32K with sharing vocabulary between source and target sides.

\subsection{Comparison Methods}
\label{sec:comparison-methods}
We compare ExtraPhrase with several existing methods to increase the training data size as follows.
We use the training set of each dataset described in Section \ref{sec:dataset} to construct pseudo data.

\paragraph{Oversampling}
This strategy is the simplest approach to increase the dataset size.
We sample source-summary pairs from the genuine training set and add the sampled instances to training data.
Thus, the training data constructed by this approach contains genuine data only.

\paragraph{Back-translation}
As described in Section \ref{sec:intro}, back-translation is the widely used approach to increase the training data in sequence-to-sequence tasks~\cite{sennrich-etal-2016-improving,edunov-etal-2018-understanding,caswell-etal-2019-tagged}.
We apply this approach to increase summarization datasets.
We train a neural encoder-decoder that generates a source text from a summary by using each training set.
Then, we input summaries in the training set to the neural encoder-decoder to generate corresponding source texts\footnote{For the back-translation approach in machine translation, we generate sentences in the source language from monolingual corpus in the target language. In the abstractive summarization, we need summaries as sentences in the target language but it is hard to obtain corpus containing summaries only. Thus, we use genuine training data.}.
We use the pairs of pseudo source texts and genuine summaries as pseudo training data.

\paragraph{Self-training}
\newcite{He2020Revisiting} indicated that self-training is effective in the abstractive summarization task.
In self-training, we train a neural encoder-decoder that generates a summary from a source text by using each training set.
Then, we input source texts in the training set to the neural encoder-decoder to generate corresponding summaries.
We use the pairs of pseudo summaries and genuine source texts as pseudo training data.

\paragraph{ExtraPhrase}
We apply ExtraPhrase to each training set.
In the headline generation, we construct pseudo summaries from the source sentence in the training data.
Because ExtraPhrase generates pseudo summary in sentence unit, the number of sentences in generated summary is not reduced in the case of the multi-sentence source text.
Thus, we use the first three sentences in the source document to reduce the number of input sentences beforehand in the document summarization.
As described in Section \ref{sec:proposed_method}, we apply ExtraPhrase to each sentence one-by-one, and then concatenate them in the original order.
In this study, we use spaCy\footnote{\url{https://spacy.io/}} \cite{Honnibal_spaCy_Industrialstrength_Natural_2020} for dependency parsing in the extractive summarization step. 
For the paraphrasing step, we use English-to-German and German-to-English translation models\footnote{\url{https://github.com/pytorch/fairseq/tree/main/examples/translation}} constructed by \newcite{ng-etal-2019-facebook}.
We translate sentences with beam width 5.

For all pseudo training data, we attach a special token, \verb|<Pseudo>|, to the front of the source text because \newcite{caswell-etal-2019-tagged} indicated that this strategy improves the performance in training on pseudo data.

\subsection{Encoder-Decoder Architecture}
\label{sec:model}
We use the de-facto standard neural encoder-decoder model, Transformer~\cite{NIPS2017_3f5ee243} in our experiments.
We use the Transformer for back-translation and self-training in addition to each abstractive summarization.
We use the Transformer (base model) setting described in \newcite{NIPS2017_3f5ee243} as our architecture.
The setting is widely used in studies on machine translation~\cite{NIPS2017_3f5ee243,ott-etal-2018-scaling}.
In detail, we use the implementation in the fairseq\footnote{\url{https://github.com/pytorch/fairseq}}~\cite{ott-etal-2019-fairseq} for our experiments.

\subsection{Results}
\label{subsec:main-result}
Tables \ref{gigaword-rouge} and \ref{cnndm-rouge} show F1 based ROUGE-1, 2, and L scores for each setting on the headline generation and document summarization tasks respectively.
These tables also show the amount of training data for each configuration.
As in these tables, we use the same size of training data for each method except for Genuine only.

Tables \ref{gigaword-rouge} and \ref{cnndm-rouge} indicate that Oversampling outperforms Genuine only.
These results indicate that the more training data we prepare, the better performance an encoder-decoder achieves even if the training data contains many duplications.
For Back-translation and Self-training, they achieve better performance than Genuine only but their scores are comparable to ones of Oversampling in both tasks.
These results imply that the improvements of their approaches are not based on the quality of their generated pseudo data, but based on the increase of training data.
Since Back-translation and Self-training require training an additional model to construct pseudo data, these approaches are more costly than Oversampling.

In contrast, our ExtraPhrase achieves better performance than Oversampling and other data construction approaches.
In particular, our pseudo training data significantly improves the ROUGE-2 score in the headline generation.
For the document summarization, our pseudo training data significantly improves all ROUGE scores\footnote{These results are statistically significant according to Student's t-test ($p < 0.05$) in comparison with Genuine only.}.
These results indicate that ExtraPhrase is more effective than existing approaches including oversampling, back-translation, and self-training to construct pseudo data for the abstractive summarization tasks.

\section{Analysis}

\begin{table*}
\centering
\begin{tabular}{lrrrrr}
\hline
Method & \#Genuine & \#Pseudo & ROUGE-1 & ROUGE-2 & ROUGE-L \\
\hline
Genuine only & 1K & -- & 4.84 & 0.58 & 4.66 \\
Oversampling & 3.8M & -- & 9.89 & 1.39 & 9.30 \\
\hline
Back-translation & 1K & 3.8M & 12.19 & 2.43 & 11.31 \\
Self-training & 1K & 3.8M & 7.27 & 1.07 & 6.98 \\
\hline
\textbf{ExtraPhrase} & 1K & 3.8M & \textbf{23.58} & \textbf{6.56} & \textbf{21.12} \\
\hline
ExtraPhrase w/o Paraphrasing & 1K & 3.8M & 22.56 & 5.25 & 19.87 \\
Extractive & -- & -- & 18.72 & 4.26 & 17.09 \\
\hline
\end{tabular}
\caption{\label{low-resource-gigaword}
F1 based ROUGE scores for the headline generation task in low-resource setting.
}
\end{table*}

\begin{table*}
\centering
\begin{tabular}{lrrrrr}
\hline
Method & \#Genuine & \#Pseudo & ROUGE-1 & ROUGE-2 & ROUGE-L \\
\hline
Genuine only & 1K & -- & 2.48 & 0.29 & 2.45 \\
Oversampling & 287K & -- & 13.63 & 0.89 & 12.63 \\
\hline
Back-translation & 1K & 286K & 9.73 & 0.50 & 8.92 \\
Self-training & 1K & 286K & 14.37 & 1.52 & 13.36 \\
\hline
\textbf{ExtraPhrase} & 1K & 286K & \textbf{34.47}& \textbf{12.91} & \textbf{31.36} \\
\hline
ExtraPhrase w/o Paraphrasing & 1K & 286K & 32.95 & 12.07 & 29.44 \\
Extractive & -- & -- & 28.52 & 8.02 & 23.83 \\
\hline
\end{tabular}
\caption{\label{low-resource-cnndm}
F1 based ROUGE scores for the document summarization task in low-resource setting.
}
\end{table*}

\subsection{Low-resource Setting}
As described in Section \ref{sec:proposed_method}, ExtraPhrase is probably robust even when the amount of genuine training data is small.
To investigate this issue, we conduct experiments in such few training data.

We randomly sample 1K source text and summary pairs from each training set in the headline generation and document summarization tasks.
Then, we conduct the same experiments in Section \ref{sec:exp} by using the sampled 1K instances as genuine training data.
In short, we construct pseudo training data from the rest of each training data and combine the pseudo data with the sampled genuine data for training.
For Self-training and Back-translation, we train neural encoder-decoders with the sampled 1K instances, and then apply them to the rest of training data for the pseudo data construction.

Tables \ref{low-resource-gigaword} and \ref{low-resource-cnndm} show the F1 based ROUGE scores of each method on the headline generation and document summarization tasks when we have a small amount of genuine training data.
These tables indicate that Oversampling outperforms Genuine only in both tasks.
Thus, duplicated training data improves the performance of encode-decoders for the abstractive summarization.
These results are consistent with the results in Section \ref{subsec:main-result}.

Table \ref{low-resource-gigaword} shows that Back-translation outperforms Oversampling in the headline generation.
Table \ref{low-resource-cnndm} shows that Self-training outperforms Oversampling in the document summarization.
These results indicate that previous approaches might be more effective than oversampling if we apply them to the appropriate task.

For ExtraPhrase, it achieves significantly better performance than others in both tasks.
Thus, ExtraPhrase is more effective when the amount of the genuine training data is small.
The lowest parts of Tables \ref{low-resource-gigaword} and \ref{low-resource-cnndm} show the results of ExtraPhrase without paraphrasing for the ablation study.
In ExtraPhrase w/o Paraphrasing setting, we train the model with genuine and pseudo training data generated by ExtraPhrase without the paraphrasing step.
Moreover, Extractive in these parts shows the ROUGE scores of summaries generated by the extractive summarization step.
These parts indicate that ExtraPhrase outperforms the one without paraphrasing.
Thus, we need the paraphrasing step to improve the quality of the pseudo training data although the setting excluding paraphrasing significantly outperforms others.
Moreover, ROUGE scores of Extractive are much lower than ones of ExtraPhrase.
This result implies that we need to train a neural encoder-decoder by using the pseudo data as the training data to generate better abstractive summaries.

\subsection{Diversity of Pseudo Summaries}
We assume that our ExtraPhrase can generate more diverse summaries in comparison with the self-training approach.
To verify this assumption, we compare pseudo summaries generated by Self-training and ExtraPhrase.

\begin{table*}
\centering
\begin{tabular}{llrr}
\hline
Task & Method & BLEU & BERTScore \\
\hline
\multirow{2}{*}{Headline generation} & Self-training & 28.64 & 92.44 \\
 & ExtraPhrase & 1.51 & 86.19 \\
\hline
\multirow{2}{*}{Document summarization} & Self-training & 19.91  & 90.02 \\
 & ExtraPhrase & 5.89 & 87.33 \\
\hline
\end{tabular}
\caption{\label{bleu-bert}
BLEU scores and F1 based BERTScores between genuine training data and pseudo training data. 
}
\end{table*}

Table \ref{bleu-bert} shows BLEU scores~\cite{Papineni02bleu:a} between genuine summaries in each training data and generated pseudo summaries.
In addition, this table shows F1 based BERTScores~\cite{Zhang2020BERTScore:} of them as the indicator of semantic similarities.
This table indicates that both BERTScores of Self-training and ExtraPhrase are remarkably high.
This result means that the generated summaries are semantically similar to genuine summaries.
Thus, generated summaries are suitable as pseudo data semantically.

In contrast, the BLEU score of ExtraPhrase is much lower than one of Self-training.
This result indicates that ExtraPhrase generates pseudo summaries that contain many different phrases from the genuine summaries in comparison with Self-training.
Therefore, ExtraPhrase can generate much more diverse summaries than Self-training.

\begin{table*}
\centering
\begin{tabular}{lrrrrrr}
\hline
Task & Ratio & Difference & ROUGE-1 & ROUGE-2 & ROUGE-L \\             
\hline
Headline generation & 0.86 & -5 & 35.14 & 15.13 & 28.59 \\
\hline
Document summarization & 0.81 & -297 & 13.76 & 1.09 & 13.07 \\
\hline
\end{tabular}
\caption{\label{rouge-BT}
F1 based ROUGE scores between source texts generated by back-translation and genuine source texts.
}
\end{table*}

\subsection{Quality of Back-translation}
As described in Section \ref{sec:intro}, the back-translation approach for the abstractive summarization task is essentially impossible because it requires restoring source texts from summaries without any additional information.
Thus, we investigate the quality of source texts generated by Back-translation.

Table \ref{rouge-BT} shows the length difference and ratio between genuine and source text generated by Back-translation.
This table indicates that the generated source texts are shorter than the original genuine data.
This result implies that Back-translation fails to restore the full information in the genuine data.
In other words, this result implies that it is difficult to generate source texts from summaries.

Table \ref{rouge-BT} also shows ROUGE scores of source texts generated by Back-translation when we regard the genuine source texts as the correct instances to investigate whether the generated texts correspond to the genuine data.
For the document summarization, ROUGE scores are extremely low.
This result also indicates that Back-translation fails to generate source texts.

On the other hand, ROUGE scores on the headline generation are much higher than ones on the document summarization.
This result implies that Back-translation might restore the core parts of source texts from summaries.
Because the headline generation is the task to generate a headline from a given sentence, the summary (headline) often contains the dominant part of the source sentence.
We consider this property causes such high scores.

\subsection{Efficiency of Pseudo-data Generation}
Our proposed ExtraPhrase does not require additional neural encoder-decoders such as the back-translation and self-training approaches.
We discuss the advantage of this property.

Table \ref{aws-cost} shows time\footnote{Consuming times are calculated in case of one GPU. } required by each pseudo data construction method.
This table also shows costs when we use Amazon EC2, which is a cloud computing service, to construct pseudo data.
This table indicates that Back-translation and Self-training require much time to train their neural encoder-decoders.
In contrast, for ExtraPhrase, we do not spend any time on such training.
Therefore, ExtraPhrase is much more cost-efficient than others as in Table \ref{aws-cost}.

\begin{table*}
\centering
\begin{tabular}{llrrr}
\hline
Task & Method & Training & Generation & Cost \\
\hline
& Back-translation & 256 H & 7 H & 333 USD  \\
Headline generation & Self-training & 256 H & 4 H & 328 USD  \\
& \textbf{ExtraPhrase} & -- & 7 H & 12 USD \\
\hline
& Back-translation & 384 H & 16 H & 511 USD \\
Document summarization & Self-training & 320 H & 8 H & 417 USD  \\
& \textbf{ExtraPhrase} & -- & 15 H & 26 USD \\
\hline
\end{tabular}
\caption{\label{aws-cost}
Cost on pseudo data generation using Amazon Elastic Compute Cloud (Amazon EC2). 
}
\end{table*}

\section{Related Work}

\paragraph{Data Augmentation} 
Back-translation and self-training are widely used techniques in data augmentation for sequence-to-sequence tasks \cite{sennrich-etal-2016-improving,kiyono-etal-2019-empirical,parida-motlicek-2019-abstract,He2020Revisiting}. 

\newcite{sennrich-etal-2016-improving} proposed back-translation to augment training data for machine translation by translating monolingual data on the target side to generate source side pseudo data. 
\newcite{edunov-etal-2018-understanding} reported the effectiveness of the back-translation approach in large-scale monolingual settings for machine translation. 
In addition, \newcite{hoang-etal-2018-iterative} introduced an iterative version by repeatedly applying back-translation several times.
\newcite{caswell-etal-2019-tagged} reported that attaching a pseudo tag to back-translated data improves the performance.
Back-translation is an effective approach for machine translation but it is unrealistic to apply the approach to abstractive summarization.

In self-training, we train a model on genuine data and apply it to generate pseudo data.
\newcite{zhang-zong-2016-exploiting} applied self-training to enlarge parallel corpus for neural machine translation.
\newcite{He2020Revisiting} introduced noisy self-training that uses dropout as the noise in while decoding in self-training.
These studies reported the effectiveness of self-training but self-training is hard to generate diverse pseudo data \cite{gu2018non}.

\paragraph{Perturbation} 
Using perturbation that is a small difference from a genuine data can be regarded as data augmentation~\cite{kobayashi-2018-contextual}.
\newcite{takase-kiyono-2021-rethinking} investigated the performance of various perturbations including adversarial perturbations~\cite{Goodfellow2015}, word dropout~\cite{NIPS2016_076a0c97}, and word replacement on various sequence-to-sequence tasks.
Since these perturbations are orthogonal to our ExtraPhrase, we can combine them with ours.
In fact, \newcite{takase-kiyono-2021-rethinking} reported that simple perturbations such as word dropout are useful on pseudo data generated by back-translation.

\paragraph{Extractive Summarization}
We use a sentence compression method in the extractive summarization step.
\newcite{dorr-etal-2003-hedge} proposed a rule-based sentence compression method which uses linguistically motivated heuristics.
\newcite{filippova-altun-2013-overcoming} proposed a supervised sentence compression method which learns important sub-trees from training data.
In addition, recent studies applied neural methods to the sentence compression task~\cite{filippova-etal-2015-sentence,kamigaito-etal-2018-higher}.
In this study, we adopt a rule-based method based on the syntax tree of the given sentence to compress the sentence because it does not require any supervised model and training corpora.

\newcite{nikolov-hahnloser-2020-abstractive} combined the sentence extraction with paraphrasing as an unsupervised summarization method.
Their approach is similar to our study in terms of using such combination but the purpose is different.
The purpose of our study is to construct pseudo training data, not to propose a novel unsupervised method.

\paragraph{Paraphrasing}
We conduct paraphrasing on extracted summaries to boost diversity in pseudo data.
\newcite{10.1007/978-3-540-27779-8_27} proposed a thesaurus-based method to paraphrase generation by substituting words with their synonyms.
\newcite{narayan-paraphrase} proposed a grammar-based method that samples paraphrase candidates from probabilistic context-free grammars learned from a training corpus.
In addition, recent studies proposed neural-based methods for paraphrase generation by formulating this problem as a sequence-to-sequence task~\cite{prakash-etal-2016-neural,10.5555/3298023.3298027}.
Moreover, previous studies used machine translation models to generate paraphrase via round-trip translation~\cite{sun-zhou-2012-joint,mallinson-etal-2017-paraphrasing}.
Since high-performance machine translation models are publicly obtainable, we apply round-trip translation as a paraphrasing method in this study.

\section{Conclusion}
This paper proposes a novel strategy, ExtraPhrase, to generate pseudo data for abstractive summarization tasks.
ExtraPhrase consists of two steps: extractive summarization and paraphrasing.
We obtain the important parts of an input by the extractive summarization, and then obtain diverse expressions by the paraphrasing.
Experimental results indicate that ExtraPhrase is more effective than other pseudo data generation methods such as back-translation and self-training.
Moreover, we show that ExtraPhrase is much more cost-efficient than others in pseudo data construction.

\bibliography{acl_latex}
\bibliographystyle{acl_natbib}

\end{document}